\documentclass[11pt]{article}
\usepackage{coling2014}
\usepackage{times}
\usepackage{url}
\usepackage{tabu}
\usepackage[utf8]{inputenc}
\usepackage{latexsym}
\usepackage{natbib}
\usepackage{linguex}
\usepackage{MnSymbol}
\usepackage{pgfplots}
\usepackage{tikz}
\usetikzlibrary{shapes}
\usetikzlibrary{positioning}

\tikzstyle{inline text}=[text height=1.5ex, text depth=0.25ex,
  yshift=0.5mm]
\tikzstyle{state}=[shape=circle,draw=blue!50,fill=blue!10]
\tikzstyle{bistate}=[shape=circle,draw=blue!50,fill=blue!10, font=\footnotesize]
\tikzstyle{lightedge}=[<-,dotted]
\tikzstyle{mainstate}=[state,thick]
\tikzstyle{endstate}=[state,thick,minimum size=1.5cm]
\tikzstyle{mainedge}=[<-,thick]

\title{Incorporating Semi-supervised Features into Discontinuous
  Easy-first Constituent Parsing
}

\author{Yannick Versley \\
  Department of Computational Linguistics \\
  University of Heidelberg\\
  {\tt versley@cl.uni-heidelberg.de}
}

\date{}

\begin{document}
\maketitle
\begin{abstract}
  This paper describes adaptations for \textsc{EaFi}, a parser for
  easy-first parsing of discontinuous constituents, to adapt it to
  multiple languages as well as make use of the unlabeled data that
  was provided as part of the SPMRL shared task 2014.
\end{abstract}

\blfootnote{This work is licenced under a Creative Commons Attribution
  4.0 International License. Page numbers and proceedings footer are
  added by the organizers. License details: \url{http://
creativecommons.org/licenses/by/4.0/}}

\section{Introduction}
The SPMRL shared task 2014 \citep{seddah14sharedtask} augments the
2013 shared task dataset -- dependency and constituent trees for
several languages, including discontinuous constituent trees for
Swedish and German -- with unlabeled data that allows for
semisupervised parsing approaches.

The following sections explain (i) how multilingual adaptation was
performed using the joint information from dependency data and
constituency data (with the help of the Universal POS tagset mapping
\citep{petrov12universalpos} where available), in section
\ref{sec:adaptation};
(ii)
the use of word clusters for semi-supervised parsing
(section \ref{sec:clusters}, and (iii) the
addition of a bigram dependency language model to the parser
(section \ref{sec:bigrams}).

\section{Easy-first parsing of discontinuous constituents}
The \textsc{EaFi} parser uses the easy-first parsing approach
of \cite{goldberg10easyfirst} for discontinuous constituent parsing.
It starts with the sequence of terminals with word forms, lemmas,
and part-of-speech tags, and progressively applies the parsing
action that has been classified as most certain. Because the
classifier only uses features from a small window around the
action, the number of feature vectors that have to be computed
and scored is linear in the number of words, contrary to approaches
that perform parsing based on a dynamic programming approach.

By using a \emph{swap} action similar to the online reordering
approach of \cite{nivre09improved}, \textsc{EaFi} is able to
perform nonprojective constituent parsing in sub-quadratic
time, with an actual time consumption being close to linear.

In order to learn the classifier for the next action, the training
component of \textsc{EaFi} runs the parsing process until the
first error (early stopping, cf.\ \citealp{collins04incremental}). The feature
vectors of the erroneous action and of the highest-scoring
action are used to perform a regularized AdaGrad update \citep{duchi11adagrad}.

\begin{figure}
\centerline{\includegraphics[width=0.8\textwidth]{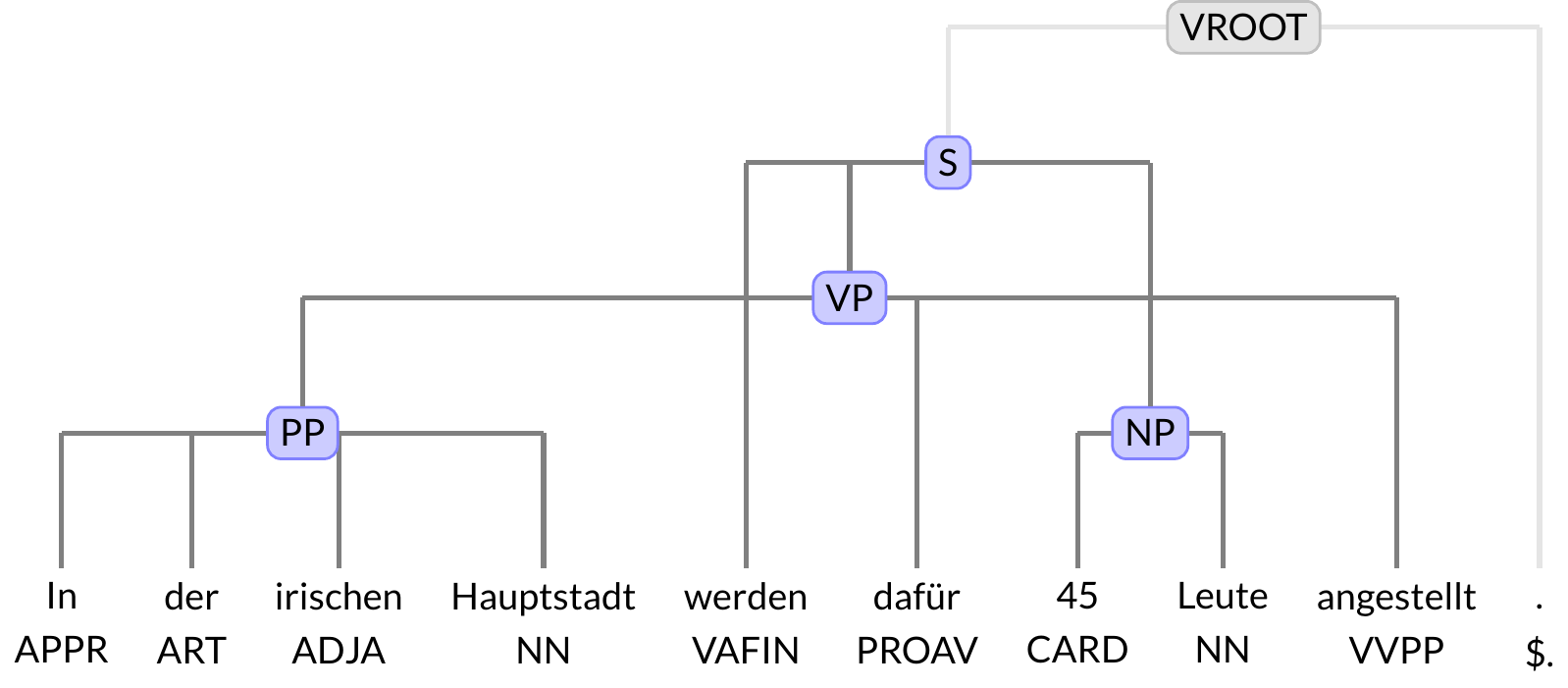}}
\centerline{``\emph{In the Irish capital, 45 people were hired for it.}''}

\caption{\label{fig:sampletree}A discontinuous constituent tree.
The \texttt{VP} node has block-degree 3.}
\end{figure}

\subsection{Multilingual Adaptations}
\label{sec:adaptation}
The EaFi parser uses two components that are language-specific and
not contained in the treebank that is used for training: the first
is a head table that is used to induce the head (pre-)terminal
of a constituent, and the second is a list of ``\emph{special}''
part-of-speech tags where the parser augments the POS values in
features with the word forms.

We induce a \textbf{head table} by interposing the dependency data
in CoNLL format with the constituency trees. Based on this, the
\emph{actual head}
of a phrase is the a preterminal that has a governor outside the
phrase. A constituent may have multiple heads when constituency and
dependency criteria do not match exactly; In cases such as
coordination constructions or appositions, the head determination
always follows the rules of the dependency scheme. The \emph{head
  constituent} is the constituent that has the head (preterminal)
as part of its yield.

From these \emph{observed} head constituents, we then try to derive a
head table in the format used by \textsc{rparse} \citep{maier10direct}
and \textsc{DiscoDop} \citep{cranenburgh12efficient} by finding a
priorization of the head constituent labels that fits the actual heads
maximally well.

For each constituent label, we start with a \emph{candidate set} of
all labels of head constituents, and try to find a priorization of
these constituent labels that fits the observed head constituents:

\begin{enumerate}
  \item Look for phrases where two daughter constituent labels from
    the candidate set occur. These are called \emph{conflicts} because
    the assigned head would depend on the order in the head rule.

    The \emph{score} of a label is the number of \emph{wins} (where
    this label and another candidate co-occur and this label has
    the actual head) minus the number of \emph{losses} (where this
    label and another candidate co-occur and the other candidate
    is the actual head constituent).

  \item The label from the candidate set with the highest score is
    appended to the rule. To decide on right-to-left or left-to-right
    precedence, look at conflicts between two instances of this label
    and count the number of conflicts that have been resolved towards
    the right/left constituent of those cases.

    Remove the label from the candidate set, and all conflicts that
    contain this candidate. If any labels remain in the candidate
    set, start again at (1.)
  \end{enumerate}

For the list of \textbf{special categories}, our intuition is that
these will be most useful in cases such as PP attachment (which
motivated their treatment as a special case in the case of
\cite{goldberg10easyfirst}, and possibly conjunctions.

We use the \emph{Universal Tagset Mapping} of
\cite{petrov12universalpos}
where it is available to make a three-way split between normal POS tags,
closed-class POS tags, and punctuation.

\begin{itemize}
  \item In the case of tagsets that have a Universal POS tag
    mapping, and tags that are mapped to \texttt{ADP}
    (adpositions) and \texttt{CONJ} (conjunctions) are included
    in the closed-class tags. Tags that have a universal POS
    mapping as \texttt{.} (punctuation) count as punctuation.
  \item In the case where no such mapping is available, we look
    at the count of types and tokens.

    If a tag has more instances containing punctuation than those
    containing containing a letter, and the number of tokens that
    contain a letter is less than 5, this tag is treated as
    punctuation.

    If a tag has more than 100 occurrences, while it only occurs with
    less than 40 different word forms, it is treated as a closed-class
    tag.
\end{itemize}

\subsection{General tuning}
EaFi uses online learning with a hash kernel to realize the learning
of parameters -- in particular, AdaGrad updates \citep{duchi11adagrad}
with
forward-backward splitting (FOBOS) for L1 regularization \citep{duchi09fobos}. Several
parameters influence the performance of the parser:

\begin{itemize}
  \item The size of the weight vector -- because a hash kernel is
    used, collisions of the hash function on the available dimension
    can have a negative impact on the performance.

    We use a 400MB weight vector, which still allows training on
    modestly-sized machines, but leaves room for more features than the 80MB
    weight vector used by \cite{versley14easyfirst}.
  \item The size of the regularization parameter. As FOBOS
    does not modify the weights between updates, smaller parameters
    seem to work better than larger ones.
    \cite{goldberg13probabilistic}
      suggests a value of $\lambda=\frac{0.05}{|D|}$, due to
    Alexandre Passos, where $|D|$ is the number of decisions per
    epoch.

    In our case, a value of $\lambda=\frac{0.001}{N}$ (for $N$ the
    number of sentences in the training set) works
    considerably better than the $\lambda=\frac{0.1}{N}$ that was
    used in the initial results reported by \cite{versley14easyfirst}.
\end{itemize}

\section{Integrating semi-supervised features}

\subsection{Using word clusters}
\label{sec:clusters}
Augmenting word forms with word clusters is univerally recognized
as a straightforward way to improve the generalization performance
of a parser. In discriminative parsers such as the dependency parser
of \cite{koo08semisupervised}, features that use surface forms are
complemented by duplicated features where the word forms are (wholly
or in part) replaced by clusters. A discriminative framework also
allows to use both clusters and reduced clusters.

\cite{candito10clusters} have shown that word clusters can productively
be incorporated into a generative parser such as the Berkeley Parser,
which uses a PCFG with latent annotations (PCFG-LA). In their case,
they augment the clusters with suffixes to improve the parser's
ability to assign the correct part-of-speech tags.

As \textsc{EaFi} uses discriminative parsing, we followed Koo et
al.\ in providing duplicates of features where word form features
are replaced by features using clusters.\footnote{Thanks to
  Djam\'e Seddah for providing providing Brown clusters for
  these languages.}

For all bigrams $m,n$ (both bigrams, and the skip bigrams $n_{-1}n_2$
and $n_0n_2$, the supervised model already includes the combinations
\[\hbox{W$m$W$n$ W$m$C$n$ C$m$W$n$ W$m$W$n$}\]
of words and the category. 

For each kind $K$ of clusters, we additionally include combination of
category and the cluster of the respective head word:
\[\hbox{C$m$K$n$ K$m$C$n$ C$m$K$m$C$n$ C$m$C$n$K$m$ C$m$K$m$C$n$K$n$}\]

We made experiments with the original clusters and with the clusters
shortened to 6 bits and 4 bits, respectively, in which the full
clusters performed best. The final model combines features using
the full clusters with features using the 6-bit cluster prefixes.

\subsection{Using a Dependency Bigram Language Model}
\label{sec:bigrams}
For models with a generative component, self-training (as in
\citealp{mcclosky06selftrain}) can provide tangible benefits. Indeed,
\cite{suzuki09semisup} show that it is possible to
reach improvements in dependency parsing beyond what is possible
with word clustering when combining a discriminative model
that uses word clusters with an ensemble of generative models
that are used as features.

While the approach of Suzuki et al.\ works with a dynamic programming
model of parsing, \cite{zhu13unlabeled} show that it is also possible to use
lexical dependency statistics learned from a large corpus to improve
a state-of-the-art shift-reduce parser for constituents.

Following Zhu et al., we add features to indicate, for the position
pairs (0,1), (1,2), (0,2), whether they belong to the top-10\%
quantile of non-zero values for one particular head word (\texttt{HI}),
to the top-30\% (\texttt{MI}), have a non-zero value (\texttt{LO}),
or a zero value (\texttt{NO}). The association scores are either
determined on the raw counts (\emph{Raw}), on proportions normalized
on the head word (\emph{L1}) or scored using the $G^2$ likelihood
ratio of \cite{dunning93statistics}.

The bigram association strength feature is taken both by itself and
paired with the POS tags of the words in question.

\section{Experiments}
Among the treebanks used in the SPMRL shared task, German and Swedish
have discontinuous constituents -- in this case, German has a large 
number of them (about ten thousand on the five thousand
sentences of the test set), while Swedish only has very few (only fifty
discontinuous phrases in the 600 sentences of the test set).

Based on prior experiments, learning on the larger German dataset
was run for 15 epochs, whereas training on the Swedish dataset
was run for 30 epochs.

\section{Results and Discussion}
Table \ref{tab:nonessential} shows how the adaptations to the purely
supervised part of \textsc{EaFi} influence the results based on
results for German gold tags. In particular, the data-driven
head table and special POS tags has a slight positive effect.
Increasing the size of the weight vector does not seem to have
strong effect, which implies that the existing weight vector is
sufficient for the feature set used in the experiments.
However, a different setting for the regularization constant yields
a rather large difference (almost +4\%), indicating that the previous
setting was suboptimal.

\begin{figure}
  \centerline{
\begin{tikzpicture}
     \begin{axis}[
         ymin = 75, 
        width  = 0.45*\textwidth,
        height = 5cm,
        ymajorgrids = true,
        ylabel = {F$_1$ German (gold), $\ell\leq 40$},
        xtick = data,
        scaled y ticks = true,
        enlarge x limits=0.1,
        legend cell align=left,
        legend style={
                at={(0.4,0.90)},
                anchor=south east,
                column sep=1ex
        }
    ]
        \addplot[style={mark=x}]
             coordinates {(5,76.47) (10, 77.07) (15, 77.42)
                (20, 77.42) (25, 77.41) (29, 77.40)};
        \addplot[style={mark=*}]
             coordinates {(5, 79.66) (10, 80.84) (15, 81.58)
                (20, 81.97) (25, 82.20) (29, 82.21)};
        \addplot[style={mark=+}]
             coordinates {(5, 79.27) (10, 79.28) (15, 78.84)
                (20, 79.58) (25, 80.54) (29, 80.48)};
         \legend{0.1, 0.01, 0.001}
    \end{axis}
     \end{tikzpicture}
\hfill
\begin{tikzpicture}
     \begin{axis}[
         ymin = 70, 
        width  = 0.45*\textwidth,
        height = 5cm,
        ymajorgrids = true,
        ylabel = {F$_1$ German (pred), $\ell\leq 40$},
        xtick = data,
        scaled y ticks = true,
        enlarge x limits=0.1,
        legend cell align=left,
        legend style={
                at={(0.4,0.90)},
                anchor=south east,
                column sep=1ex
        }
    ]
        \addplot[style={mark=x}]
             coordinates {(5,73.09) (10, 73.60) (15, 73.82)
                (20, 74.00) (25, 74.04) (29, 74.06)};
        \addplot[style={mark=*}]
             coordinates {(5, 76.83) (10, 77.85) (15, 78.61)
                (20, 78.78) (25, 79.08) (29, 79.28)};
        \addplot[style={mark=+}]
             coordinates {(5, 76.45) (10, 78.72) (15, 79.12)
                (20, 79.43) (25, 79.85) (29, 79.97)};
         \legend{0.1, 0.01, 0.001}
    \end{axis}
\end{tikzpicture}
     }
  \caption{\label{fig:l1reg}Influence of the regularizer on
    learning performance (yield $F_1$ versus epochs)}
\end{figure}
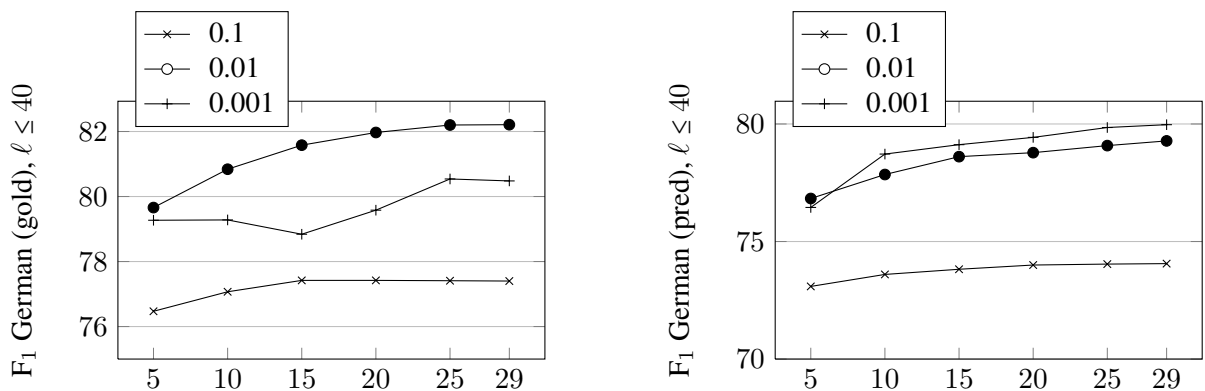

\begin{table}
\centerline{%
  \begin{tabular}{l|ccc|ccc|}
    &$F_1$&EX&UAS&NP&PP&VP\\
    \hline
    original \textsc{EaFi}&76.35&41.35&87.02&75.1&83.2&57.9\\
    + multilingual adaptation&76.70&41.93&87.12&75.5&83.1&58.5\\
    + 400MB weights &76.65&41.79&87.11&75.4&83.3&58.6\\
    + l1=0.001&80.60&48.43&89.75&79.8&86.9&65.0\\
    \hline
    \end{tabular}}
\caption{\label{tab:nonessential}Parser parameters (German only, dev
  set, gold preprocessing, $\ell \leq 70$)}
\end{table}

In tables \ref{tab:semisup} and \ref{tab:semisup_pred}, we find the
supervised initial results together with experiments regarding the
use of clusters and their granularity, and the use of features based
on the bigram language model.

Both for Swedish and for German, we see that adding cluster-based
features improves the results considerably, with an increase of
+1.3\% in the case of German and of slightly more than +3.5\% in Swedish
for predicted tags and +1.7\% and +2.6\%, respectively, for gold tags.

We also see that the shortest version of the clusters (4bit) works
less well than the others, while clusters shortened to 6-bit prefixes
are relatively close to the results using full clusters.

\begin{table}
  \centerline{\small%
    \begin{tabular}{l|ccc|ccc|ccc}
    &$F_1$&EX&UAS&NP&PP&VP\\
    \hline
    \emph{German}
    supervised      &80.60&48.43&89.75&79.8&86.9&65.0\\
    clusters (full) &\textbf{82.45}&\textbf{50.65}&\textbf{90.54}&81.3&87.9&67.4\\
    clusters (6bit) &80.56&48.07&89.66&80.1&87.2&62.7\\
    clusters (4bit) &80.81&47.95&89.83&79.9&87.1&64.2\\
    clusters (full+6bit)&
                     81.64&49.73&90.05&81.1&87.5&65.3\\
                     \hline
    clust(full+6bit)+Bigram(\emph{Raw})&
                     80.20&49.43&89.66&81.1&87.3&59.9\\
    clust(full+6bit)+Bigram(\emph{L1})&
                     80.60&49.09&89.62&81.3&87.1&60.4\\
    clust(full+6bit)+Bigram(\emph{LL})&
                     \textbf{82.20}&\textbf{50.86}&\textbf{90.31}&81.2&87.8&65.9\\
    \hline
    \hline
    \emph{Swedish}
    supervised     &73.13&18.90&74.72&76.6&66.7&64.6\\
    clusters (full)&\textbf{75.77}&\textbf{22.97}&\textbf{76.93}&78.3&71.1&66.0\\
    clusters (6bit)&75.25&22.76&76.69&78.0&70.7&66.4\\
    clusters (4bit)&74.98&21.95&76.17&78.0&70.7&66.6\\
    clusters (full+6bit)&
                    76.02&24.39&76.30&78.8&70.4&66.5\\
                    \hline
    clust(full+6bit)+Bigram(\emph{Raw})&
                    75.93&22.36&\textbf{77.18}&79.1&69.8&64.7\\
    clust(full+6bit)+Bigram(\emph{L1})&
                    76.12&\textbf{22.76}&77.08&78.9&71.2&66.8\\
    clust(full+6bit)+Bigram(\emph{LL})&
                    \textbf{76.37}&22.56&77.16&79.2&71.6&68.5\\
    \hline
   \end{tabular}}
  \caption{\label{tab:semisup}Integration of semisupervised features
    (gold preprocessing)}
\end{table}

\begin{table}
  \centerline{\small%
    \begin{tabular}{l|ccc|ccc|ccc}
    &$F_1$&EX&UAS&NP&PP&VP\\
    \hline
    \emph{German}
    supervised     &78.63&44.97&87.51&77.4&85.9&59.3\\
    clusters (full)&\textbf{79.96}&\textbf{47.35}&\textbf{88.16}&79.1&86.7&61.2\\
    clusters (6bit)&79.34&45.91&87.69&78.1&86.4&60.8\\
    clusters (4bit)
                   &78.59&44.53&87.45&77.2&85.6&59.6\\
    clusters (full+6bit)
                   &79.77&46.79&87.96&79.1&86.4&60.9\\
    \hline
    \hline
    clust(full+6bit)+Bigram(\emph{Raw})
                   &79.95&47.09&88.14&79.1&86.7&61.5\\
    clust(full+6bit)+Bigram(\emph{L1})
                   &\textbf{80.07}&\textbf{47.25}&88.12&79.2&86.7&61.8\\
    clust(full+6bit)+Bigram(\emph{LL})
                   &79.96&47.19&\textbf{88.20}&79.3&87.0&60.8\\
    \hline
    \emph{Swedish}
    supervised     &70.72&16.06&73.32&74.9&66.1&61.5\\
    clusters (full)&74.26&18.90&75.78&77.0&69.7&65.1\\
    clusters (6bit)&74.05&19.72&75.57&77.0&69.6&65.2\\
    clusters (4bit)&72.06&17.89&74.50&76.1&66.9&62.9\\
    clusters (full+6bit)&
                   \textbf{74.39}&\textbf{20.33}&\textbf{76.06}&76.5&69.0&66.4\\
    \hline
    clust(full+6bit)+Bigram(\emph{Raw})&
                   \textbf{74.46}&19.92&\textbf{76.24}&77.4&70.2&66.2\\
    clust(full+6bit)+Bigram(\emph{L1})&
                   74.19&20.12&75.82&77.1&70.4&65.7\\
    clust(full+6bit)+Bigram(\emph{LL})&
                    74.39&\textbf{20.33}&76.06&76.5&69.0&66.4\\
    \hline
   \end{tabular}}
  \caption{\label{tab:semisup_pred}Integration of semisupervised features
    (predicted tags\&morph)}
\end{table}

\begin{table}
  \centerline{%
    \begin{tabular}{l|ccc|ccc|}
    &$F_1$&EX&UAS&NP&PP&VP\\
    \hline
    \multicolumn{7}{l|}{\emph{German, gold preprocessing}}\\
supervised $\diamondsuit$
                    &73.53&38.47&85.54&74.5&82.8&56.1\\
clusters (full+6bit)&75.76&39.71&86.51&76.1&83.7&60.3\\
+Bigram (LL) $\heartsuit$
                    &76.46&41.05&86.94&76.4&84.4&60.5\\[0.4ex]
\hline
    \multicolumn7{l|}{\emph{German, predicted tags+morph}}\\
supervised $\diamondsuit$
                    &71.96&35.12&83.15&71.7&80.9&55.0\\
clusters (full+6bit)&73.35&35.98&83.80&73.1&82.1&55.4\\
+Bigram (LL) $\heartsuit$
                    &73.90&37.00&84.16&73.7&82.6&56.7\\
\hline
\multicolumn7{l|}{\emph{Swedish, gold preprocessing}}\\
supervised $\diamondsuit$
                    &81.16&31.73&80.81&83.2&79.4&77.0\\
clusters (full+6bit)&82.10&34.89&81.56&84.6&80.3&77.1\\
+Bigram(LL) $\heartsuit$
                    &82.49&34.89&81.37&84.7&79.2&79.2\\
\hline
\multicolumn7{l|}{\emph{Swedish, predicted tags+morph}}\\
supervised $\diamondsuit$
                    &79.17&28.42&79.63&82.6&76.1&73.1\\
clusters (full+6bit)&80.61&30.83&79.71&83.4&78.2&76.6\\
+Bigram(LL) $\heartsuit$
                    &80.58&30.68&80.14&83.6&78.5&76.7\\
\hline
\end{tabular}}
  \caption{\label{tab:official}Results on the test set. Results from
    the official submission are marked with $\diamondsuit$ (supervised
    run) and $\heartsuit$ (semi-supervised run).}
\end{table}

\section{Conclusions}
In this paper, we have reported adaptations with the dual goal of,
firstly, using the \textsc{EaFi} engine for parsing multiple languages by
harnessing existing dependency conversions and tagset mappings to
provide head rules and lists of closed-class tags; secondly, of
improving on these supervised learning results by incorporating
features based on data from large corpora without manual annotation,
namely Brown clusters and a dependency bigram language model.

Experimental results show that these two improvements are well-suited
to improve the capabilities of the parser. At the same time, they
demonstrate that techniques that are well-known in dependency parsing
can also be harnessed to create parsers for discontinuous constituent
structures that work better than existing parsers that are based on
treebank LCFRS grammars, making it a practical solution for the
parsing of discontinuous structures such as extraposition and scrambling.

\bibliographystyle{acl}
\bibliography{ml,anaphora}
\end{document}